\pdfoutput=1

\documentclass[11pt]{article}

\usepackage[]{ACL2023}

\usepackage{times}
\usepackage{latexsym}
\usepackage{enumitem}
\usepackage{hyperref}
\usepackage{url}
\usepackage{multirow}
\usepackage{bbm}
\usepackage{graphicx}
\usepackage{amssymb}
\usepackage{amsmath} 
\usepackage{booktabs}
\usepackage{lscape}
\usepackage{pifont}
\usepackage{graphicx}
\usepackage{caption}
\usepackage{subcaption}
\usepackage{CJKutf8}
\usepackage{cleveref}

\usepackage[T1]{fontenc}

\usepackage[utf8]{inputenc}

\usepackage{microtype}

\usepackage{inconsolata}

\usepackage[ruled,vlined]{algorithm2e}

\newcommand\methodname{\textcolor{black}{\textsc{EZSwitch}}}
\newcommand\benchmarkname{\textcolor{black}{\textsc{CSPref}}}

\title{Linguistics Theory Meets LLM: Code-Switched Text Generation via Equivalence Constrained Large Language Models}

\author{Garry Kuwanto$^1$, Chaitanya Agarwal$^2$, Genta Indra Winata$^{\dagger3}$\thanks{\hspace{1.7mm}The work was done outside the affiliation. $^{\dagger}$The authors are senior authors.}\text{ }\hspace{0.5mm}, Derry Tanti Wijaya$^{\dagger1,4}$ \\
  $^1$Boston University $\quad$ $^2$Deccan AI $\quad$ $^3$Capital One $\quad$ $^4$Monash University Indonesia \\
  \texttt{gkuwanto@bu.edu, chaitanya@deccan.ai}\\
  \texttt{genta.winata@capitalone.com, derry.wijaya@monash.edu}
  }

\begin{document}
\maketitle
\begin{abstract}
Code-switching, the phenomenon of alternating between two or more languages in a single conversation, presents unique challenges for Natural Language Processing (NLP). Most existing research focuses on either syntactic constraints or neural generation, with few efforts to integrate linguistic theory with large language models (LLMs) for generating natural code-switched text. In this paper, we introduce $\methodname$, a novel framework that combines Equivalence Constraint Theory (ECT) with LLMs to produce linguistically valid and fluent code-switched text. We evaluate our method using both human judgments and automatic metrics, demonstrating a significant improvement in the quality of generated code-switching sentences compared to baseline LLMs. To address the lack of suitable evaluation metrics, we conduct a comprehensive correlation study of various automatic metrics against human scores, revealing that current metrics often fail to capture the nuanced fluency of code-switched text. Additionally, we create $\benchmarkname$, a human preference dataset based on human ratings and analyze model performance across ``hard'' and ``easy'' examples. Our findings indicate that incorporating linguistic constraints into LLMs leads to more robust and human-aligned generation, paving the way for scalable code-switching text generation across diverse language pairs.\footnote{The code for $\methodname$ is released at \url{https://github.com/gkuwanto/ezswitch} and $\benchmarkname$ is released at \url{https://huggingface.co/datasets/garrykuwanto/cspref}.}
\end{abstract}

\section{Introduction}






%

\begin{figure}[!t]
    \centering
    \includegraphics[width=\linewidth]{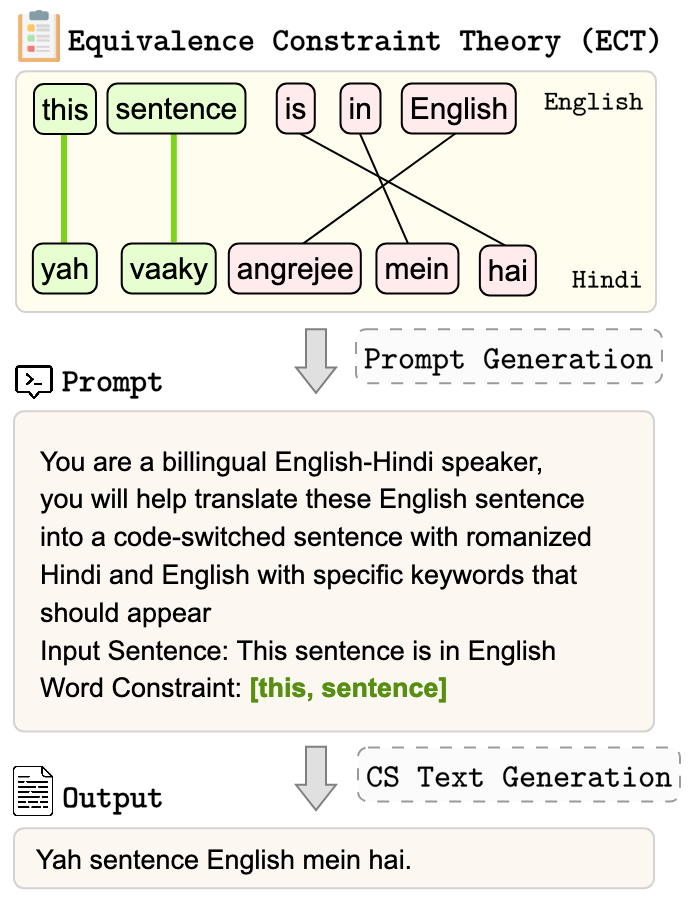}
    \caption{Example of $\methodname$. The top panel shows the word-level alignment between English and Hindi. The middle panel displays the input to the LLM, including the original English sentence and word constraints derived from ECT. The bottom panel shows the resulting code-switched output sentence generated by the LLM.}
    \label{fig:ect}
\end{figure}

Code-switching, the practice of alternating between languages within a sentence or discourse, is prevalent in multilingual communities~\cite{myers1997duelling}. This phenomenon poses significant and long-standing challenges for natural language processing (NLP) tasks, such as natural language understanding, text generation, and speech processing. Large Language Models (LLMs) often struggle to effectively handle the complexity and variability inherent in code-switched language, resulting in suboptimal performance in these areas~\cite{winata2021multilingual,zhang2023multilingual}. Linguistic studies have shown that code-switching does not occur randomly but follows specific patterns or grammatical rules~\cite{poplack1980sometimes,myers1994social}. Leveraging this linguistic information helps predicting where code-switching naturally occurs. However, many of these linguistic constraints require complex grammatical structures  as inputs, such as those from external word aligners or constituency parsers that can be difficult for models to process. In addition, constituency parsers are not always available and may introduce errors on  distant languages~\cite{winata2019code}.

To address the complexity of learning code-switching patterns in NLP, existing approaches often treat code-switched language as a distinct language, developing new neural machine translation (NMT) models trained on non-code-mixed parallel data to generate code-switched data~\cite{winata2019code,gupta2021training}. These models learn in a supervised manner, eliminating the need for grammar parsers that might introduce errors. While these methods have shown promising results, they come with substantial computational costs and resource demands. Training a new NMT model from scratch to generate code-switched data is both expensive and time-consuming, making it impractical for resource-constrained environments or low-resource language pairs. Additionally, the robustness of NMT models can be compromised given the scarcity of the required parallel to code-switched training data.

One of the critical challenges in generating code-switched text is identifying valid switching points that adhere to the syntactic and semantic rules of the languages involved. The equivalence constraint theory (ECT)~\cite{poplack1980sometimes} suggests that code-switching is permissible only at points where the grammatical structures of the languages align. However, applying this theory in practice is complex, particularly when the goal is to generate text that is both fluent and contextually appropriate. To bridge the gap, we propose to leverage linguistics constraints and incorporate them seamlessly on LLMs. Our contributions can be summarized in three-fold:
\begin{enumerate}
    \item We propose $\methodname$, a code-switching generation approach that effectively integrates ECT into LLM, generating code-switched text that is both syntactically and semantically valid. An example of $\methodname$ is illustrated in \Cref{fig:ect}.
    \item We conduct a comprehensive high-quality evaluation of the generated samples via human annotation with native speakers. Given the lack of suitable automatic metrics for code-switching fluency, we employ human evaluations to assess fluency and accuracy, and analyze correlations with established metric.
    \item We release $\benchmarkname$, a pairwise preference dataset designed to assess the fluency and accuracy of code-switched generated text. This dataset holds potential value for future research in preference tuning.
\end{enumerate}
We release our code-switched data with human preference annotations to facilitate code-switching future research and evaluation.

\section{Methodology}
In this section, we provide an overview of linguistic theory and description of $\methodname$.
\subsection{Equivalence Constraint Theory (ECT)}
\label{sec:ect}
ECT posits that code-switching occurs only when it does not violate the syntactic rules of either language~\cite{poplack1980sometimes}. \citet{winata2019code} applies ECT by simplifying sentences in terms of a linear grammatical structure and allowing lexical substitution on \textit{non-crossing alignments} between parallel sentences (e.g., lexical substitution between "sentence" and "vaaky" in Figure \ref{fig:ect}). Denoting $L_1$ as the source language and $L_2$ as the target language, given a sentence in $L_1$ comprising an array of words $u_t = {a_1, a_2, ..., a_m}$ and a corresponding sentence in $L_2$ comprising an array of words $v_t = {b_1, b_2, ..., b_m}$, the alignment between $a_i$ and $b_i$ does not satisfy the constraint if there exists a pair $a_j$ and $b_j$ such that ($a_i < a_j$ and $b_i > b_j$) or ($a_i > a_j$ and $b_i < b_j$). If a switch occurs at this point, it alters the grammatical order in both languages, rendering the switch unacceptable. During the generation step, we permit \textit{any} switches that do not violate this constraint.

\subsection{Relaxed ECT}
While adhering to the core principles of Equivalence Theory \cite{poplack1980sometimes}, our approach applies and builds on the \textit{relaxed} version of ECT introduced by \citet{winata2019code} (\Cref{sec:ect}) to the task of code-switched sentence generation.  
This relaxation allows for greater flexibility in identifying potential switching points, accommodating the complexities of real-world code-switching patterns while maintaining grammatical coherence. Our implementation expands on the linear grammatical structure and non-crossing alignment criteria outlined in \Cref{sec:ect}, introducing additional flexibility to capture a broader range of code-switching phenomena. In the following section, we outline our approach to implementing the relaxed ECT for identifying switching points, developing code-switched sentence generation techniques, and establishing a comprehensive evaluation framework.

\subsection{$\methodname$}
Building on the theoretical foundation established earlier, 
our implementation of ECT to generate code-switched sentence 
comprises of three main stages: 1) obtaining translations, 2) bitext alignment, and 3) code-switched sentence generation.

\subsubsection{Obtaining Translations}

We employ two distinct methods for obtaining translations:

\paragraph{Human Translations.} We utilize existing parallel datasets that contain human-generated translations. Specifically, for Hindi we use the HinGE dataset~\citep{srivastava2021hinge}, and for Tamil and Malayalam we use the Samanantar \citep{ramesh2022samanantar} dataset. These datasets contain human-translated parallel sentences, providing a reliable \textit{gold} standard for translations. 

\paragraph{LLM Translations.} For LLM translations, we utilize Llama3 8B, a compact and open-source large language model. The translation process involves directly prompting the model. This approach allows us to generate translations using a consistent method across all input sentences and provides the \textit{silver} standard for translations. 

\subsubsection{Bitext Alignment}
After obtaining translations, we use the GIZA++ tool \cite{och2003systematic} to align the sentences. We then apply the resulting alignments alongside the Equivalent Constraint Theory, as described in \Cref{sec:ect}, to generate a list of potential switching points.

\subsubsection{Code-Switched Sentence Generation}

Once we identify switching points using the ECT approach, we guide language models to generate code-switched sentences by inputting the original monolingual sentence into the model. To construct the prompt effectively, we incorporate the following information:

\paragraph{Bilingual Context Setting.} We instruct the language model to act as a bilingual speaker of the two languages involved (e.g., Hindi and English).

\paragraph{Exemplar Provision.} We provide an example of a code-switched sentence along with its monolingual counterpart to demonstrate the desired style of output.

\paragraph{Constraint Application.} We specify key words from both languages, derived from the valid switching points (\Cref{sec:ect}) that should appear in the code-switched output.

\paragraph{Generation.} The model generates a code-switched sentence based on the given context, example, constraints, and input sentence. We employ three different prompting methods in our experiments, each applied bidirectionally: A $\rightarrow$ CS (A and B), and B $\rightarrow$ CS (B and A), where the former is from language A to a code-switched text of A and B, and the latter is from language B to a code-switched text of B and A.

\paragraph{$\methodname$.} We use switching points identified with ECT using the silver translations generated by Llama3 8B. The model is given specific words from these LLM-generated translations to include in the code-switched output. This bidirectional approach across the three methods allows us to comprehensively explore code-switching patterns initiated from both languages, providing a robust comparison of the different prompting strategies.

\section{Experimental Setup}
\subsection{Models}
We evaluate our approach using three different LLMs: Aya23~\cite{aryabumi2024aya}, Llama3 8B and Llama3.1 8B~\cite{dubey2024llama}. \citet{aryabumi2024aya} introduce Aya 23 8B, an open-weight large language model specifically designed for multilingual tasks. It is trained on a diverse corpus of languages, making it particularly suitable for code-switching applications. \citet{dubey2024llama} develop Llama3 8B, a compact yet powerful language model from the Llama3 series. With 8 billion parameters, it offers a balance between computational efficiency and performance, making it ideal for exploring code-switching in resource-constrained settings. Llama3.1 8B is an improved version of Llama3 8B. This model features refined training techniques and an updated dataset, incorporating the latest advancements in language modeling and potentially offering enhanced performance in multilingual and code-switching tasks. All experiments are run on a single NVIDIA L40 GPU with 48GB of memory.

\subsection{Translation Directions}
We consider two directions of code-switched generation in our experiments: English to Code-Switched and Indic Language (Hindi, Tamil, Malayalam) to Code-Switched. These directions apply to each language pair in our study, allowing us to examine code-switching patterns initiated from both languages.

\subsection{Baselines} 
We do not use ECT-identified switching points. Instead, we simply instruct the model to create a code-switched version of the input sentence without specifying particular words to include. As another comparison, we We use switching points identified with ECT using the gold (human) translations from the parallel datasets (HinGE for Hindi, Samanantar for Tamil and Malayalam), we called it \textbf{Human ECT}. The model is given specific words from these human translations to include in the code-switched output.

\subsection{Datasets}
\label{sec:dataset}
Our experiments utilize three parallel corpora, each corresponding to a distinct language pair, as detailed in \Cref{tab:datasets}. For the human translations employed in our Human ECT method, we rely on the parallel sentences available in these datasets. In contrast, for the LLM translations used in $\methodname$, we generate translations using the Llama3 8B model.

\begin{table}[!t]
    \centering
    \resizebox{.49\textwidth}{!}{
    \begin{tabular}{llc}
        \toprule
        \textbf{Language Pair}        & \textbf{Source}       & \textbf{Size} \\ 
        \midrule
        Hindi-English (hi-en)            & HinGE                & 2,766          \\
        Tamil-English (ta-en)            & Samanantar (WAT 2020)& 2,000          \\
        Malayalam-English (ml-en)        & Samanantar (WAT 2020)& 2,000          \\
        \bottomrule
    \end{tabular}
    }
    \caption{Summary of datasets used in the experiments, including source and size.}
    \label{tab:datasets}
\end{table}

\subsection{Evaluation}
In our evaluation, we assess the effectiveness of various automatic metrics in capturing human judgment scores of fluency and accuracy of the code-switched text. Specifically, we experiment with COMET \citep{rei2020comet}, permuting the reference language to gauge sensitivity across different reference settings. Additionally, we employ GPT-4o-mini as an evaluator to explore leveraging large language models (LLMs) for assessing code-switching, given the subjectivity inherent in human evaluations. This approach allows us to compare traditional automatic metrics with LLM-based evaluation in terms of alignment with human preferences.

\paragraph{COMET.} We compute COMET scores by using English (L1) and the Indic language (L2) as the reference language in separate runs, referred to as COMET\_l1 and COMET\_l2. To obtain a more comprehensive evaluation, we also take the average of these two scores (COMET\_avg). This permutation helps in evaluating how robust COMET is across different reference configurations, capturing variability in code-switched translations. Among the metrics, comet\_avg demonstrates the highest correlation with human evaluations, making it a more reliable option for code-switching assessment.
\paragraph{GPT4 Evaluation.} In addition to classical metrics, we conduct an evaluation using GPT4 as a proxy for human judgment. We provide GPT4 (specifically, \textsc{gpt-4o-mini}) with the exact same set of instructions as our human evaluators, asking it to rate the generated sentences for both accuracy and fluency. This setup allows us to directly compare the correlation between GPT4 ratings and human annotations, providing insight into the potential of large language models as evaluators.

\subsection{Human Evaluation}
We perform human evaluations of the code-switched sentences with native speakers who are bilingual and use code-switching in their daily communications. We adopt a setup inspired by~\citet{kuwanto2024mitigating} and reformulate the evaluation to focus on rating sentences rather than expressing a preference. Human evaluators, contractually employed by DeccanAI (previously, SoulAI) are asked to score the accuracy and fluency of the code-switched sentences. DeccanAI ethically recruits human evaluators in India, and evaluates them to determine their English and native language proficiency via their crowdsourcing platform. They also onboard evaluators, train them with the annotation guidelines, and obtain the annotation scores through their annotation platform.

This on 150 sample of inputs from dataset described in \cref{sec:dataset} with 18 different generation settings. Totaling 2700 sentence to rate for each language. In total, we conduct 24,300 human evaluations in total. For each code-switched sentence, we ask 3 unique evaluators to score the accuracy and fluency of the sentence on a discrete scale from 1 (lowest) to 3 (highest). While evaluating, the evaluators can see (1) the English sentence, (2) the sentence in the respective native language, and (3) the LLM generated code-switched sentence.

\section{Results}

\subsection{Automatic Metrics}
We evaluate the entire dataset described in \Cref{tab:datasets}, each of the input pairs yielded 18 different generations from the different method and model. We use GPT-4o-mini, as traditional metrics often fail to align with human judgment for code-switched text. GPT-4o-mini was used to assess both Accuracy (GPT4o$_a$) and Fluency (GPT4o$_f$) for all model and method combinations. The results are split into translations from English input and Indic input to allow for a detailed comparison of each method's performance across different language directions. Word Replacement (WR) refers to the method proposed in \citet{winata2019code}, which we use for comparison against our proposed method.
\begin{table}[!t]
    \centering
    \resizebox{0.96\linewidth}{!}{\begin{tabular}{llrr}
        \toprule
        \textbf{Method} & \textbf{Model} & \textbf{GPT4o$_a$} & \textbf{GPT4o$_f$}\\
        \midrule
        \multicolumn{4}{c}{From English Input}\\
        \midrule
        \multirow[t]{3}{*}{Baseline} & WR & 1.362 & 1.374 \\
        & Aya23 & 1.469 & 1.486 \\
        & Llama3 & 1.375 & 1.390 \\
        & Llama3.1 & 1.430 & 1.452\\
        \midrule
        \multirow[t]{3}{*}{Human ECT} & Aya23 & 1.446 & 1.449\\
        & Llama3 & 1.354 & 1.364\\
        & Llama3.1 & \textbf{1.498} & \textbf{1.512}\\
        \midrule
        \multirow[t]{3}{*}{$\methodname$} & Aya23 & 1.377 & 1.472 \\
        & Llama3 & 1.336 & 1.345 \\
        & Llama3.1 & 1.477 & 1.485 \\
        \midrule
        \multicolumn{4}{c}{From Indic Input}\\
        \midrule
        \multirow[t]{3}{*}{Baseline} & WR & 1.410 & 1.412 \\
        & Aya23 & 1.486 & 1.848 \\
        & Llama3 & 1.732 & 1.697\\
        & Llama3.1 & 1.834 & 1.823\\
        \midrule
        \multirow[t]{3}{*}{Human ECT} & Aya23 & 1.525 & 1.775 \\
        & Llama3 & 1.451 & 1.445 \\
        & Llama3.1 & \textbf{1.850} & \textbf{1.852} \\
        \midrule
        \multirow[t]{3}{*}{$\methodname$} & Aya23 & 1.528 & 1.788 \\
        & Llama3 & 1.460 & 1.450 \\
        & Llama3.1 & 1.843 & 1.849 \\
        \bottomrule
    \end{tabular}
    }
    \caption{GPT-4 evaluation of Accuracy (GPT4o$_a$) and Fluency (GPT4o$_f$) for English-to-Indic and Indic-to-English translations across different models and methods. WR refers to the word replacement method from \citet{winata2019code}. Best scores for each column within a language group are highlighted in \textbf{bold}.}
    \label{tab:my_label}
\end{table}

\subsection{Human Evaluation Results}
\label{sec:human_result}
We performed human evaluations on a sample of the entire dataset, using a random sample of 150 input sentences per Indic language. Given that we tested 3 methods, 3 models, and 2 translation directions (i.e., English $\leftrightarrow$ Indic), this resulted in a total of 2,700 ratings per Indic language. \Cref{tab:metric_mean} shows the result of each of three primary methods: \textbf{Baseline}, \textbf{Human ECT}, and \textbf{$\methodname$}, evaluated across three models: Aya23, Llama3, and Llama3.1. We use human evaluation of Accuracy and Fluency ratings as our primary evaluation criteria, alongside automatic metrics such as COMET and GPT4-based scores (denoted as GPT4o$_a$ for Accuracy and GPT4o$_f$ for Fluency). The table is divided into two main sections based on the direction of translation: \textbf{From English Input} and \textbf{From Indic Input}, allowing us to compare the performance across different language directions. Each method and model combination is evaluated to identify which setup achieves the highest fluency and accuracy as judged by human evaluators. Additionally, COMET is included as a traditional automatic metric for comparison, while GPT4-based evaluations provide an alternative automated approach to assessing fluency and accuracy using large language models.

\begin{table*}[!th]
    \centering
    \resizebox{\textwidth}{!}{
    \begin{tabular}{llrrrrr}
        \toprule
        \textbf{Method} & \textbf{Model} & \textbf{Human Accuracy} & \textbf{Human Fluency} & \textbf{COMET} & \textbf{GPT4o$_a$} & \textbf{GPT4o$_f$} \\
        \midrule
        \multicolumn{7}{c}{From English Input}\\
        \midrule
        \multirow[t]{3}{*}{Baseline} & Aya23 & 1.394 & 1.375 & \textbf{0.512} & 1.482 & 1.502 \\
         & Llama3 & 1.314 & 1.313 & 0.442 & 1.384 & 1.418 \\
         & Llama3.1 & 1.404 & 1.400 & 0.481 & 1.496 & 1.529 \\
        \cmidrule{1-7}
        \multirow[t]{3}{*}{Human ECT} & Aya23 & 1.332 & 1.377 & 0.483 & 1.458 & \textbf{1.584} \\
         & Llama3 & 1.306 & 1.301 & 0.463 & 1.396 & 1.456 \\
         & Llama3.1 & 1.421 & 1.400 & 0.499 & \textbf{1.527} & 1.573 \\
        \cmidrule{1-7}
        \multirow[t]{3}{*}{$\methodname$} & Aya23 & 1.313 & 1.369 & 0.473 & 1.418 & 1.522 \\
         & Llama3 & 1.327 & 1.285 & 0.451 & 1.396 & 1.458 \\
         & Llama3.1 & \textbf{1.424} & \textbf{1.417} & 0.491 & 1.504 & 1.556 \\
        \midrule
        \multicolumn{7}{c}{From Indic Input}\\
        \midrule
        \multirow[t]{3}{*}{Baseline} & Aya23 & 1.470 & 1.567 & 0.555 & 1.389 & 1.680 \\
         & Llama3 & \textbf{2.231} & \textbf{2.141} & 0.415 & 1.562 & 1.587 \\
         & Llama3.1 & 2.004 & 1.926 & 0.436 & \textbf{1.731} & 1.738 \\
        \cmidrule{1-7}
        \multirow[t]{3}{*}{Human ECT} & Aya23 & 1.583 & 1.613 & \textbf{0.567} & 1.433 & 1.656 \\
         & Llama3 & 1.801 & 1.711 & 0.428 & 1.393 & 1.442 \\
         & Llama3.1 & 1.888 & 1.837 & 0.513 & 1.702 & 1.782 \\
        \cmidrule{1-7}
        \multirow[t]{3}{*}{$\methodname$} & Aya23 & 1.538 & 1.605 & 0.566 & 1.438 & 1.644 \\
         & Llama3 & 1.756 & 1.691 & 0.435 & 1.384 & 1.413 \\
         & Llama3.1 & 1.904 & 1.833 & 0.510 & 1.713 & \textbf{1.791} \\
        \bottomrule
    \end{tabular}
    }
    \caption{Mean scores of Human Accuracy, Human Fluency, COMET (comet\_avg), and GPT4-based evaluations (GPT4o$_a$ for Accuracy and GPT4o$_f$ for Fluency) for Baseline, Human ECT, and $\methodname$ methods across three models (Aya23, Llama3, and Llama3.1) with English and Indic as source languages. Scores are grouped by translation direction (English-Indic and Indic-English) to highlight the performance differences based on input language. Best scores for each column within a language group are highlighted in \textbf{bold}.}
    \label{tab:metric_mean}
\end{table*}

\subsection{Correlation Between Automatic and Human Metrics}
This subsection explores the relationship between various automatic evaluation metrics and human judgments for both Accuracy and Fluency in code-switched text generation. Understanding these correlations is crucial for determining the reliability of automatic metrics as proxies for human evaluation. \Cref{tab:metric_corr} presents the Kendall’s Tau correlation scores between human ratings and several automatic metrics, including BLEU, COMET, and BERTScore, along with GPT4-based evaluations. Higher correlation values indicate a stronger alignment with human ratings, suggesting the suitability of a metric for capturing fluency and syntactic coherence in code-switched outputs. This experiment was done on the same 150 subset of input pairs from \Cref{sec:human_result}, which yielded 2,700 samples for each language, and we measure the correlations of these metrics with the average human ratings.

\subsection{Pairwise Preference Dataset}

We construct $\benchmarkname$, a pairwise preference dataset using human ratings to evaluate the performance of different models in code-switched text generation. Each pair consists of two generated code-switched sentences compared based on their human-evaluated accuracy and fluency scores. To further analyze the performance, we split the dataset into \textbf{easy} and \textbf{hard} subsets. The \textbf{easy} subset includes pairs where the difference in human ratings is high (indicating a clear preference for one generated sentence), while the \textbf{hard} subset consists of pairs with minimal differences (indicating ambiguous preferences). \Cref{tab:pairwise_stats} provides the statistics for the pairwise dataset across three languages: Hindi, Tamil, and Malayalam. We report the total number of pairs, as well as the breakdown into ``easy'' and ``hard'' subsets for each language pair.

\begin{table}[!t]
    \centering
    \resizebox{0.99\linewidth}{!}{
        \begin{tabular}{lcc}
        \toprule
         & \textbf{Accuracy} & \textbf{Fluency} \\
        \midrule
        Human Accuracy & 1.000 & 0.768 \\
        Human Fluency & 0.768 & 1.000 \\
        GPT4o$_a$ & 0.558 & 0.504 \\
        GPT4o$_f$ & 0.540 & 0.514 \\
        COMET\_avg & 0.246 & 0.290 \\
        COMET\_l1 & 0.237 & 0.285 \\
        BLEU$^*$ & 0.229 & 0.201 \\
        COMET\_l2 & 0.175 & 0.202 \\
        BERTScore\_l2\_f1 & 0.121 & 0.204 \\
        BERTScore\_l2\_recall & 0.119 & 0.200 \\
        BERTScore\_l2\_precision & 0.117 & 0.199 \\
        BERTScore\_l1\_recall & 0.073 & 0.126 \\
        BERTScore\_l1\_f1 & 0.071 & 0.128 \\
        BERTScore\_l1\_precision & 0.064 & 0.120 \\
        \bottomrule
    \end{tabular}
    }
    \caption{Kendall’s tau correlation scores between different automatic metrics and human evaluations for Accuracy and Fluency. $^*$BLEU score can only be calculated for Hindi-English as there is no code-switched references for other language pairs.}
    \label{tab:metric_corr}
\end{table}

\begin{table}[!th]
    \centering
    \resizebox{\linewidth}{!}{
        \begin{tabular}{lrrr}
            \toprule
            \textbf{Language Pair} & \textbf{Total} & \textbf{Easy} & \textbf{Hard} \\ \midrule
            Hindi-English (hi-en)             & 17,460 & 9,621 & 7,839 \\
            Tamil-English (ta-en)           & 5,034 & 4,506 & 528 \\
            Malayalam-English (ml-en)        & 8,664 & 7,517 & 1,147 \\ \bottomrule
        \end{tabular}
    }
    \caption{Statistics of $\benchmarkname$ for three languages (Hindi, Tamil, and Malayalam). ``Easy'' pairs are defined as those with high rating differences, while ``Hard'' pairs are defined as those with low rating differences.}
    \label{tab:pairwise_stats}
\end{table}

\section{Discussion}

\subsection{Model Performance}
Our experimental results, as presented in Table \ref{tab:my_label}, demonstrate that even small, open-source models like Llama3 8B can produce high-quality code-switched sentences when guided by linguistic constraints. The ECT-guided generation outperformed the baseline in both fluency and grammatical accuracy, illustrating that linguistic theory can effectively enhance the outputs of language models. Notably, Llama3.1 8B consistently achieved higher accuracy and fluency scores compared to other models, highlighting the importance of the language model's architecture and parameterization in code-switching generation.

\subsection{Method Effectiveness}
The three generation methods (Baseline, Human ECT, and $\methodname$) offer valuable insights based on human evaluation. $\methodname$ outperforms Human ECT for translations from English input, demonstrating better fluency and syntactic coherence. This suggests that $\methodname$ excels at handling English-to-Indic code-switching. For from Indic input, human evaluations show that $\methodname$ and Human ECT perform similarly, but both underperform compared to the Baseline. This may be due to the abundance of data where Indic languages act as the matrix language and English as the embedded language, making LLMs more adept at handling this type of code-switching, as they already possess knowledge of such patterns. When assessed using automatic metrics, Human ECT performed best across all measures, highlighting its ability to capture linguistic nuances that LLM-generated translations might miss. However, $\methodname$ still showed significant improvement over the Baseline, demonstrating that ECT-guided approaches can be applied effectively in low-resource scenarios. These automatic metric results underscore the need for metrics that better reflect the complexities of code-switching.

\subsection{Directional Asymmetry}

An unexpected finding was the directional asymmetry in model performance between different translation directions. Specifically, translations from Indic languages to English exhibited significantly higher fluency and accuracy than translations in the reverse direction. This asymmetry may be attributed to the training data distribution, where code-switching from Indic languages to English is more prevalent in real-world usage, making the model more familiar with this direction. Additionally, all annotators being native speakers of Indic languages and L2 learners of English might introduce bias. Their proficiency in their native language allows for more rigorous scrutiny of translations into Indic languages, potentially leading to stricter evaluations of fluency and accuracy. Conversely, translations from Indic languages to English might be rated more leniently, as English is a second language for the annotators, making them less critical of subtle errors. This directional asymmetry could also reflect real-world code-switching patterns that favor English as the matrix language, with Indic languages filling lexical gaps. Future research could explore how this asymmetry relates to linguistic theories, such as the Matrix Language Frame model, and examine the impact of annotators' bilingual proficiency on these evaluations.

\subsection{Evaluation Challenges and Human Preference Correlation}
Evaluating code-switching fluency and accuracy is challenging due to the limitations of existing automatic metrics. Based on Table \ref{tab:metric_corr}, while SentenceBLEU reduces reliance on human-generated references, it falls short in measuring fluency. Embedding-based metrics like BERTScore and COMET, although better at capturing semantic similarity, exhibit weak correlations with human evaluations, around 0.2. In contrast, GPT-4o-mini, when used as an evaluator with structured guidance, achieves a higher correlation (Kendall’s tau of 0.5). This indicates that LLMs can effectively approximate human judgment, albeit with potential stylistic biases. These findings underscore the need for specialized code-switching metrics that integrate syntax, semantics, and context, emphasizing naturalness and linguistic validity over traditional n-gram-based matching. Developing tailored metrics is essential for accurately assessing multilingual text generation, as conventional tools fail to capture the complexities inherent in code-switching.

\section{Related Work}
Code-switching has been extensively studied from both linguistic and computational perspectives. Early linguistic theories, such as ECT \citep{poplack1980sometimes}, establishes foundational principles for understanding syntactic boundaries in code-switching. Similarly, research by \citet{joshi1982processing} and \citet{pfaff1979constraints} examine structural constraints and sentence processing in bilingual contexts. Recent computational approaches have adapted these theories into neural models. For instance, \citet{winata2019code} utilized ECT to generate synthetic data for training language models, while \citet{gupta2020semi} employed pre-trained models to create code-switched text without explicit constraints. \citet{pratapa2021comparing} utilized ECT to synthetically generate code-switched text by using the Dependency Tree. And \citet{gupta2021training} adopted a Machine Translation approach to the problem. Comprehensive survey by \citet{sitaram2019survey,winata2023decades} outline the computational challenges and advancements in code-switching research.

Evaluation benchmarks, such as LinCE \citep{aguilar2020lince} and GLUECoS \citep{khanuja2020gluecos} have standardized model assessments across diverse tasks. Recent studies have also investigated automatic metrics for code-switching \citep{guzman2017metrics} and explored the use of LLMs in understanding code-switched text~\citep{de2024code}, and also generating \citep{yong2023prompting}. In this context, our work builds on these foundations by integrating linguistic constraints into LLM-based generation, addressing existing limitations in fluency and accuracy evaluation.

\section{Future Work}
\begin{figure}[!t]
    \centering
    \includegraphics[width=0.82\linewidth]{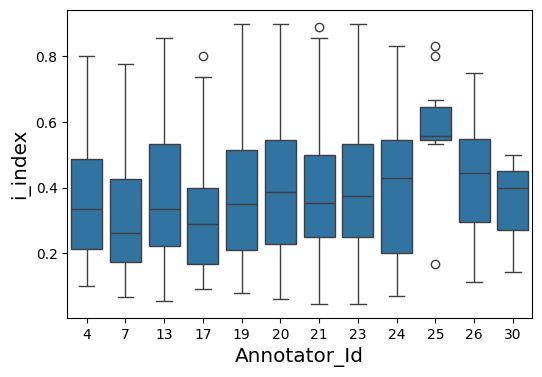}
    \caption{Distribution of I-index across annotators, representing the probability of code-switching at any given token. The I-index indicates the proportion of switch points relative to language-dependent tokens in the corpus. The variability in switching preferences among annotators highlights the individual differences in their judgment of fluency, suggesting that demographic factors may play a role in code-switching evaluation.}
    \label{fig:i_index}
\end{figure}

In our preliminary analysis of code-switching metrics, using methods like those proposed by \citet{guzman2017metrics}, we find significant variability in how annotators judged sentences, with different distributions of metrics across individuals. This variability is further highlighted by the I-index distribution for all annotators, as shown in \Cref{fig:i_index}. These findings suggest that individual preferences and linguistic backgrounds play a critical role in the evaluation of code-switching fluency. Future work should delve into the demographic data of annotators, exploring how factors like age, region, and language proficiency influence code-switching preferences. A more detailed study on how these demographic factors affect evaluation consistency could improve the robustness of human-centered code-switching metrics. In addition to exploring demographic influences, another promising direction is training LLMs to align with human preferences, conditioned on demographic data. By using our human evaluation dataset, we could develop personalized code-switching models that adapt to individual linguistic preferences and switching habits. Conditioning on demographics like age, region, and language proficiency, this approach could create a more tailored and context-aware code-switching LLM, ultimately improving the fluency and relevance of code-switched text generation for diverse users. It is also worthwhile to explore the dialectical variations in code-switching languages that exhibit different language styles~\cite{aji2022one,winata2024worldcuisines}. Developing a new metric aligned with human preferences for natural-sounding code-switching text would greatly benefit future code-switching research~\cite{winata2024metametrics,winata2024preference}.

\section{Conclusion}
We propose $\methodname$, a method for generating high-quality code-switched text by integrating Equivalence Constraint Theory (ECT) with large language models (LLMs). Our approach addresses a significant gap in code-switching generation by combining linguistic theoretical constraints with modern language models. Through a series of experiments using both human evaluations and automatic metrics, we demonstrate that our method produces more fluent and accurate code-switching sentences compared to baseline LLMs and previous works. Furthermore, we constructed a Human Preference dataset to capture human preferences to help future work with aligning to codeswitching. Our correlation analysis of automatic metrics, such as COMET, reveals that existing metrics are insufficient for evaluating code-switching text, and alternative approaches such as GPT-based evaluation achieve better alignment with human judgments.

\section*{Limitations}
Our study is intentionally limited in scope, focusing exclusively on the evaluation of open-source language models to ensure reproducibility and transparency in our findings. By concentrating on these models, we aim to establish a solid foundation for future research that can be easily replicated by other scholars in the field. Looking ahead, we plan to extend our methodology to include commercial language models, which may provide enhanced capabilities and performance. This expansion will enable us to explore the potential benefits and challenges of integrating proprietary models into our framework. Additionally, we are considering the inclusion of more languages in future studies to further enrich our research and broaden its applicability across diverse linguistic contexts.

\section*{Ethics Statement}
All aspects of this research were reviewed and approved by the Institutional Review Board of our organization. Data collection was conducted by DeccanAI. We compensate human evaluators INR 110 for every 18 sentences they evaluate, which typically takes around 20 minutes. This results in an effective pay rate of INR 330 per hour. The human evaluators work entirely remotely and interact with DeccanAI through their web platform. All evaluators are native speakers of the respective Indic languages they assess and are proficient in English. Their language proficiency is evaluated through custom online tests. Most evaluators come from major cities in India where these native languages are spoken and frequently engage in code-switched dialogues. DeccanAI provides training for the evaluators to ensure they are well-calibrated with the annotation guidelines.


\bibliography{custom}
\bibliographystyle{acl_natbib}

\appendix

\section{Switching Point Algorithm}
The algorithm for the process of getting valid switching points is as described in Algorithm \ref{alg:switchingpoints} 

\begin{algorithm}[!ht]
\SetAlgoLined
\KwResult{List of valid switching points}
\SetKwFunction{FGetValidSwitchingPoints}{GetValidSwitchingPoints}
\FGetValidSwitchingPoints{pairs}\\
{
    valid\_pairs $\gets$ [ ]\;
    \For{$i\gets1$ \KwTo $\text{length}(pairs)$}{
        valid $\gets$ true\;
        \For{$j\gets1$ \KwTo $\text{length}(pairs)$}{
            $(a_i, b_i) \gets pairs[i]$\;
            $(a_j, b_j) \gets pairs[j]$\;
            \If{$(a_i < a_j \text{ and } b_i > b_j)$ \textnormal{\textbf{or}} $(a_i > a_j \text{ and } b_i < b_j)$}{
                valid $\gets$ false\;
                \textbf{break}\;
            }
        }
        \If{valid}{
            Append $pairs[i]$ to valid\_pairs\;
        }
    }
    \Return valid\_pairs\;
}
\caption{Identification of Valid Switching Points}
\label{alg:switchingpoints}
\end{algorithm}

\section{Statistical Analysis of Human Evaluation}

\subsection{Overall Effects}

We conducted a series of one-way ANOVAs to investigate the impact of three primary factors—\textbf{Model}, \textbf{Method}, and \textbf{Direction}—on both Accuracy and Fluency scores. The results, presented in \Cref{tab:anova_results}, indicate that all factors have a statistically significant effect on both measures (all p < .001). 

Specifically, the \textbf{Model} factor shows a strong effect on Accuracy ($F = 43.71, p < .001$) and a moderate effect on Fluency ($F = 18.82, p < .001$), suggesting that different models handle the nuances of code-switched text differently. The \textbf{Method} factor also significantly influences both Accuracy ($F = 9.66, p < .001$) and Fluency ($F = 20.80, p < .001$), highlighting the importance of incorporating linguistic constraints during generation. Finally, the \textbf{Direction} factor has the highest impact on both Accuracy ($F = 323.13, p < .001$) and Fluency ($F = 293.31, p < .001$), indicating a substantial difference in translation quality between English-Indic and Indic-English directions. These findings underscore the necessity of considering all three factors when evaluating code-switching generation methods.

\begin{table}[!th]
\centering
\resizebox{0.49\textwidth}{!}{
\label{tab:anova_results}
\begin{tabular}{lrrrr}
\toprule
\textbf{Factor} & \multicolumn{2}{c}{\textbf{Accuracy}} & \multicolumn{2}{c}{\textbf{Fluency}} \\
 & F-score & p-value & F-score & p-value \\
\midrule
Model & 43.71 & <.001 & 18.82 & <.001 \\
Method & 9.66 & <.001 & 20.80 & <.001 \\
Direction & 323.13 & <.001 & 293.31 & <.001 \\
\bottomrule
\end{tabular}
}
\caption{Results of one-way ANOVAs for examining the effects of Model, Method, and Direction on Accuracy and Fluency scores. Each factor shows a significant impact on both measures (all p < .001), indicating that the choice of model, the generation method, and the translation direction significantly influence the overall quality of code-switched text generation.}
\label{tab:anova}
\end{table}
\subsection{Model Comparison}
For the model comparisons, we evaluate the performance of Llama3, Llama3.1, and Aya23 on English-Indic and Indic-English translation tasks, focusing on both Accuracy and Fluency. The detailed results for English-Indic translation are presented in \Cref{tab:model}. For accuracy, Llama3.1 significantly outperforms both Aya23 and Llama3, while no significant difference is observed between Aya23 and Llama3. In terms of fluency, Llama3 outperforms Aya23, and Llama3.1 outperforms both Llama3 and Aya23, though the difference between Aya23 and Llama3.1 is not statistically significant.

For the Indic-English translation, the comparisons show that both Llama3 and Llama3.1 significantly outperform Aya23 in terms of accuracy, but there is no significant difference between Llama3 and Llama3.1. For fluency, both Llama3 and Llama3.1 again outperform Aya23, with a slight significant difference between Llama3.1 and Llama3, as shown in \Cref{tab:model}.

\begin{table}[!th] 
    \centering
    \resizebox{0.47\textwidth}{!}{
    \begin{tabular}{llrr}
        \toprule
        \textbf{Model Comparison} & \textbf{Measure} & \textbf{Mean Diff.} & \textbf{p-value} \\ \midrule
        \multicolumn{4}{l}{English to Indic Translation} \\ \midrule
        Llama3.1 vs. Aya23         & Accuracy         & 0.1881                  & < .001           \\
        Llama3.1 vs. Llama3        & Accuracy         & 0.1859                  & .0001            \\
        Aya23 vs. Llama3           & Accuracy         & 0.0022                  & .9986            \\
        Llama3 vs. Aya23           & Fluency          & 0.1267                  & .0034            \\
        Llama3.1 vs. Llama3        & Fluency          & 0.1859                  & < .001           \\
        Aya23 vs. Llama3.1         & Fluency          & 0.0593                  & .2813            \\ \midrule
        \multicolumn{4}{l}{Indic to English Translation} \\ \midrule
        Llama3 vs. Aya23           & Accuracy         & 0.3363                  & < .001           \\
        Llama3.1 vs. Aya23         & Accuracy         & 0.4326                  & < .001           \\
        Llama3.1 vs. Llama3        & Accuracy         & 0.0963                  & .0807            \\
        Llama3 vs. Aya23           & Fluency          & 0.1889                  & < .001           \\
        Llama3.1 vs. Aya23         & Fluency          & 0.2881                  & < .001           \\
        Llama3.1 vs. Llama3        & Fluency          & 0.0993                  & .0452            \\ \bottomrule
    \end{tabular}
    }
    \caption{Tukey’s post-hoc test results for model comparisons across English-Indic and Indic-English translation directions. The table reports the mean differences and p-values for Accuracy and Fluency between Aya23, Llama3, and Llama3.1 models, highlighting significant variations in performance depending on the model used.}
    \label{tab:model}
\end{table}

\subsection{Method Comparisons}
The method comparisons analyze Human ECT, baseline approaches, and $\methodname$ for English-Indic and Indic-English translation. \Cref{tab:method} presents the results. For English to Indic translation, Human ECT significantly outperforms Baseline in terms of accuracy, but the difference between Baseline and $\methodname$, as well as between Human ECT and $\methodname$, is not significant. In terms of fluency, both Human ECT and $\methodname$ outperform the Baseline significantly, with no significant difference observed between Human and $\methodname$.

For Indic to English translation, Human ECT and $\methodname$ both outperform the Baseline for accuracy, and there is no significant difference between Human ECT and $\methodname$. Similarly, for fluency, both methods show significant improvements over Baseline, but no substantial difference between Human and $\methodname$ is observed, as shown in \Cref{tab:method}
\begin{table}[t]
    \centering
    \resizebox{.49\textwidth}{!}{
    \begin{tabular}{llrr}
        \toprule
        \textbf{Method Comparison} & \textbf{Measure} & \textbf{Mean Diff.} & \textbf{p-value} \\ \midrule
        \multicolumn{4}{c}{\textbf{English to Indic Translation}} \\ \midrule
        Human ECT vs. Baseline      & Accuracy         & 0.1348                  & .0058            \\
        Baseline vs. $\methodname$ & Accuracy         & 0.0993                  & .0597            \\
        Human ECT vs. $\methodname$ & Accuracy        & 0.0356                  & .6939            \\
        Human ECT vs. Baseline      & Fluency          & 0.1578                  & .0002            \\
        $\methodname$ vs. Baseline & Fluency          & 0.1319                  & .0022            \\
        Human ECT vs. $\methodname$ & Fluency         & 0.0259                  & .7843            \\ \midrule
        \multicolumn{4}{c}{\textbf{Indic to English Translation}} \\ \midrule
        Human ECT vs. Baseline      & Accuracy         & 0.1281                  & .0157            \\
        $\methodname$ vs. Baseline & Accuracy         & 0.1496                  & .0036            \\
        Human ECT vs. $\methodname$ & Accuracy        & 0.0215                  & .8881            \\
        Human ECT vs. Baseline      & Fluency          & 0.2022                  & < .001           \\
        $\methodname$ vs. Baseline & Fluency          & 0.1763                  & .0001            \\
        Human ECT vs. $\methodname$ & Fluency         & 0.0259                  & .8101            \\ \bottomrule
    \end{tabular}
    }
    \caption{Tukey’s post-hoc test results for method comparisons across English-Indic and Indic-English translation directions. The table presents the mean differences and p-values for Human ECT, $\methodname$, and Baseline methods, indicating significant variations in Accuracy and Fluency depending on the generation approach used.}
    \label{tab:method}
\end{table}

\subsection{Direction Comparison}

We also compare the direction of translation (English-Indic vs. Indic-English) to determine if there are asymmetries in model performance. The results in \Cref{tab:direction} indicate that Indic to English translation significantly outperforms English to Indic translation for both Accuracy and Fluency, with mean differences of 0.4684 and 0.4037, respectively (p < .001 for both measures).

\begin{table}[!th]
    \centering
    \begin{tabular}{@{}lcc@{}}
        \toprule
        \textbf{Measure} & \textbf{Mean Difference} & \textbf{p-value} \\ \midrule
        Accuracy         & 0.4684                  & < .001           \\
        Fluency          & 0.4037                  & < .001           \\ \bottomrule
    \end{tabular}
    \caption{Tukey’s post-hoc test results for mean differences and p-values in Accuracy and Fluency between Indic-to-English and English-to-Indic translation directions. Significant differences indicate that the direction of translation has a substantial impact on performance, with Indic-to-English outperforming English-to-Indic in both metrics.}
    \label{tab:direction}
\end{table}

\section{Prompt Details}
In \cref{tab:prompts} we present the specific prompts used across different methods in our code-switching experiments. The Translate prompt is used when we generate the translations to get alignment. The Baseline prompt is used when we evaluate LLM to generate codeswitching. The ECT prompt is used for both $\methodname$ and Human ECT. The GPT Eval prompt defines the structure for evaluating code-switching output based on accuracy and fluency.
\begin{table*}[ht]
    \centering
    \begin{tabular}{ll}
        \hline
        \textbf{Method} & \textbf{Prompt} \\
        \hline
        Translate &
        \begin{tabular}[c]{l}
        Translate the following \texttt{lang1} sentence to \texttt{lang2}:\\
        \texttt{<Input Sentence>}\\
        \end{tabular} \\
        \hline
        Baseline &
        \begin{tabular}[c]{l}
        You are a Bilingual \texttt{lang1}-\texttt{lang2} speaker, 
        you will help translate these \texttt{lang1}\\
        sentences  
        into a code-mixed sentence with Romanized 
        
        \texttt{lang2} and \texttt{lang1}\\
        \texttt{<Input Sentence>}\\
        \end{tabular} \\
        \hline
        \methodname &
        \begin{tabular}[c]{l}
        You are a Bilingual \texttt{lang1}-\texttt{lang2} speaker, 
        you will help translate these \texttt{lang1} \\
        sentences
        into a code-mixed sentence with Romanized 
        \texttt{lang2} and \texttt{lang1} \\with specific keywords that 
        should to appear.\\
        \texttt{<Input Sentence>}\\
        Words wanted: \texttt{<List of Words>}\\
        \end{tabular} \\
        \hline
        GPT Eval &
        \begin{tabular}[c]{l}
        
You are provided with triplets of sentences. The first two sentence in each triplet \\
is the original monolingual sentences. The second sentence is a generated \\
code-switched sentence. Your task is to evaluate the generated sentence based \\
on two criteria: Accuracy and Fluency. You will score each criterion on a scale \\
from 1 to 3, where 1 is the lowest and 3 is the highest.
When evaluating the \\
generated sentences, focus on the content and meaning. Ignore any extra \\
formatting, alignment artifacts, or additional explanatory text. Judge the \\
sentence to determine its accuracy and fluency. \\
original\_l1: \texttt{<Original Lang1 Sentence>}\\
original\_l2: \texttt{<Original Lang2 Sentence>}\\
generated: \texttt{<Code Switched Sentence>}\\
        \end{tabular} \\
        \hline
    \end{tabular}
    \caption{Prompts used in our experiment.}
    \label{tab:prompts}
\end{table*}

\section{Human Evaluation}
\subsection{Annotation Guidelines}
The following guidelines are provided to human evaluators to assess the model's responses. Evaluators rate the generated sentences based on two criteria: \textbf{Accuracy} and \textbf{Fluency}. The original sentence is in English, Indian local languages (Hindi, Malayalam, and Tamil), and evaluators must adhere to the rubrics outlined below.

\subsubsection{General Guidelines}
\begin{itemize}
    \item \textbf{MUST}: Be objective while rating the responses.
    \item \textbf{MUST}: Strictly follow the rubrics for Accuracy and Fluency evaluation.
    \item Score each criterion on a scale from 1 to 3, where 1 is the lowest and 3 is the highest.
    \item Ignore formatting, and any additional explanatory text generated by the language model. Only focus on meaning and context.
    \item If the model fails to generate a response, assign a score of 1 for both Accuracy and Fluency.
\end{itemize}

\subsubsection{Accuracy}
Accuracy measures how well the generated sentence preserves the meaning and information of the original sentence and whether the code-switched terms are used correctly. The scores are as follows:
\begin{itemize}
    \item \textbf{1. Low Accuracy}: 
    \begin{itemize}
        \item Significant deviations from the original meaning.
        \item Key information is missing, altered, or repeated redundantly.
        \item Code-switched terms are incorrect or inappropriate.
        \item Introduces new information not present in the original sentence.
        \item Key details are altered or repeated redundantly.
    \end{itemize}
    \item \textbf{2. Moderate Accuracy}:
    \begin{itemize}
        \item Minor deviations from the original meaning.
        \item Most key information is present but may have slight errors.
        \item Most code-switched terms are appropriate but with minor mistakes.
    \end{itemize}
    \item \textbf{3. High Accuracy}:
    \begin{itemize}
        \item Preserves the original meaning fully.
        \item All key information is present and correct.
        \item Code-switched terms are accurate and appropriately used.
    \end{itemize}
\end{itemize}

\subsubsection{Fluency}
Fluency measures how natural and easy to understand the generated sentence is, considering grammar, syntax, and the smooth integration of code-switching. The scores are as follows:
\begin{itemize}
    \item \textbf{1. Low Fluency}: 
    \begin{itemize}
        \item The sentence is difficult to understand or awkward.
        \item Poor grammar or syntax in either language.
        \item Code-switching disrupts the flow of the sentence.
    \end{itemize}
    \item \textbf{2. Moderate Fluency}: 
    \begin{itemize}
        \item The sentence is understandable but may have awkward or unnatural phrasing.
        \item Acceptable grammar and syntax in both languages.
        \item Code-switching is somewhat smooth but not perfectly integrated.
    \end{itemize}
    \item \textbf{3. High Fluency}: 
    \begin{itemize}
        \item The sentence is natural and easy to understand.
        \item Good grammar and syntax in both languages.
        \item Code-switching is smooth and seamless, enhancing the sentence flow.
    \end{itemize}
\end{itemize}

\subsection{Inter Annotator Agreement}
As seen in \Cref{tab:human_eval_agreement}, the inter-annotator agreement, measured by Krippendorff’s alpha, reveals varying levels of consensus across languages, with the highest agreement for Hindi. While Fluency is generally lower, this is expected as Fluency is more of a subjective measure.

Throughout the evaluation process, we continuously monitored the quality of annotations by measuring inter-annotator agreement at regular intervals. If the agreement metric indicated significant divergence in scores, particularly when individual annotators' ratings deviated notably from the group consensus, we conducted alignment meetings. These meetings were used to clarify the guidelines and ensure a consistent understanding of the evaluation criteria among the annotators. During these sessions, any inconsistencies were discussed and resolved to improve consistency, especially in subjective aspects like Fluency. This iterative process helped ensure the reliability of the final evaluations and minimized discrepancies in the ratings.

\begin{table}[!ht]
    \centering
    \begin{tabular}{lcc}
        \toprule
        \textbf{Language} & \textbf{Fluency} & \textbf{Accuracy} \\ 
        \midrule
        Tamil             & 0.321                                     & 0.445                                      \\
        Malayalam         & 0.405                                     & 0.423                                      \\
        Hindi             & 0.646                                     & 0.720                                      \\
        \bottomrule
    \end{tabular}
    \caption{Krippendorff's alpha scores for inter-annotator agreement on Fluency and Accuracy across Tamil, Malayalam, and Hindi.}
    \label{tab:human_eval_agreement}
\end{table}

\end{document}